\documentclass{ecai}
\usepackage{times}
\usepackage{graphicx}
\usepackage{latexsym}
\usepackage[ruled,vlined]{algorithm2e}
\usepackage{amssymb}
\usepackage{amsmath}
\usepackage{lipsum}
\usepackage{hyperref}
\usepackage{multirow}

\begin{document}

\title{Automatic Bridge Bidding Using \\Deep Reinforcement Learning}

\author{Chih-Kuan Yeh\institute{Department of Electrical Engineering, National Taiwan University, email: \url{b01901163@ntu.edu.tw}} \and Hsuan-Tien Lin\institute{Department of Computer Science and Information Engineering, National Taiwan University, email: \url{htlin@csie.ntu.edu.tw}}}

\maketitle

\begin{abstract}

  Bridge is among the zero-sum games for which artificial intelligence has not yet outperformed expert human players. The main difficulty lies in the bidding phase of bridge, which requires cooperative decision making under partial information. Existing artificial intelligence systems for bridge bidding rely on and are thus restricted by human-designed bidding systems or features. In this work, we propose a pioneering bridge bidding system without the aid of human domain knowledge. The system is based on a novel deep reinforcement learning model, which extracts sophisticated features and learns to bid automatically based on raw card data. The model includes an upper-confidence-bound algorithm and additional techniques to achieve a balance between exploration and exploitation. Our experiments validate the promising performance of our proposed model. In particular, the model advances from having no knowledge about bidding to achieving superior performance when compared with a champion-winning computer bridge program that implements a human-designed bidding system.
  
\end{abstract}

\section{Introduction}
\label{sec:intro}

Games have always been a challenging testbed for artificial intelligence (AI).
Even for games with simple and well-defined rulesets, AI often needs to follow highly complex strategies to gain victory.
One set of works on game AI focuses on full information games including chess, go, and Othello~\cite{silver2016mastering}, whereas the other set studies incomplete information games such as poker and bridge~\cite{ho2015contract,sandholm2010state,yakovenko2016poker}.
In both cases, traditional works usually excel by embedding the knowledge of the best human players as computable strategies; however, researchers have recently shifted their focus to machine learning, allowing AI players to develop effective strategies automatically from data~\cite{ho2015contract,silver2016mastering,yakovenko2016poker}.
	
Bridge, a standard 52-card game that requires players to be both cooperative and competitive, is one of the most appraised partial-information games for humans and for AI. The four players of the bridge game are commonly referred to as North, East, West and South, and form two opposing teams (North-South and East-West). Each team aims to achieve the highest score in a zero-sum scenario.
	
A single bridge game starts with a deal followed by two phases: bidding and playing. A deal distributes 13 random cards to each player, and the cards are hidden from other players---each player only sees partial information about the deal. The bidding phase runs an auction to determine the declarer of the contract, where the contract affects the score that the declarer's team can get in the playing phase. The auction proceeds around the table in a clockwise manner, where each player chooses from one of the following actions: PASS, increasing the current value of the bid with respect to an ordered set of calls $\{ 1\clubsuit, 1\diamondsuit, 1 \heartsuit, 1 \spadesuit, 1NT, 2\clubsuit ,..., 7NT\}$, DOUBLING and REDOUBLING. The first two actions are general ones for deciding the contract, while the latter two are special, less-used actions that modify the scoring function for the playing phase. The bidding sequence ends when three consecutive PASSes are placed, and the last bid becomes the final contract. The number in the final contract (such as $4$ in $4\spadesuit$) plus $6$ represents the number of rounds that the team aims to win in the playing phase to achieve the contract (commonly referred to as ``make''), and the symbol (such as~$\spadesuit$) reflects the trump suit in the playing phase.

In the playing phase of the bridge game, there are $13$ rounds where each player shows one card from her/his hand and compares the values of the cards based on some rulesets with the trump suit having some priority. The player with the highest-valued card among the four is the winner of the round. After the $13$ rounds, the score of the declarer's team is calculated by a lookup table based on the final contract and the number of winning rounds of the declarer's team, where making the contract leads to a positive score for the declarer's team, and not making (failing) the contract results in a positive score for the the opponent's team.

Bidding is an understandably a difficult task because of the incomplete-information setting. Given that each player can only see 13 out of 52 cards, it is impossible for a single player to infer the best contract for her/his team. Thus, each bid in the bidding phase needs to serve as a suggestion towards an optimal contract, information-exchanging between team members, or both. That is, a good bidding strategy should balance between exploration (exchanging information) and exploitation (deciding an optimal contract). Nevertheless, because the bid value needs to be monotonically increasing during the auction, the information exchanging is constrained to the extent that it does not exceed the optimal contract. It is also possible that the two opposing teams may both try to exchange information during the bidding phase, called bidding with competition, which blocks the other team's information-exchanging opportunities.

In real human bridge games, the best human players are often indistinguishable in terms of their professional competence in the playing phase. Thus, their competence in the bidding phase is the primary game-deciding factor. The abovementioned facts indicate the relative difficulty of the bidding phase over the playing phase for human players. The difficulty holds true in the case of designing AI players as well.
For the playing phase, it has been shown that AI players are competitive against professional human ones. For example, in 1998, the GIB program finished in the 12th place among 35 professional human players in a no-bidding bridge contest \cite{ginsberg1999gib}. Nevertheless, for the bidding phase, most existing AI players are based on replicating human-designed rules of bidding, commonly referred to as human bidding systems~\cite{carley2004program,lindelof1983cobra,stanier1975bribip,wasserman1970realization}. The replication generally makes AI players less competitive to human players in the bidding phase, as explained below.
	
One of the main difficulties in replicating a human bidding system is the inevitable ambiguity of the bids. Human bidding systems are designed to have rules that cover different situations, but the rules can be overlapping. Therefore, based on the cards of one player and the other players' bids, there can be conflicting suggestions that can be arrived at from different rules, with every suggestion being a legitimate bid under the system. Human players are expected to resolve this ambiguity intelligently and select an appropriate choice from the conflicting suggestions; in addition, professional human players devote a considerable amount of time to practice together with team members to reduce the ambiguity through mutual understanding. When AI players try to replicate human bidding systems, it is extremely challenging to reach the same level of mutual understanding that human players can achieve to resolve the ambiguity, making AI players inferior in the bidding phase.

The ambiguity of the bids arises primarily because human bidding systems need to be simple enough to be memorizable by human players. Thus, the rules within the systems are often simple, too. On the other hand, if there were a bidding system for AI players instead of human players, the rules may not need to be so simple, and the ambiguity issue may be resolved to improve the performance of AI players in the bidding phase.

The aforementioned ideas were the motivation behind some existing works on enhancing AI for bridge bidding. Some works begin by considering a human bidding system and then resolve the ambiguity using different techniques. For instance, combining lookahead search with human bidding system was studied by Gamb{\"a}ck et al.~\cite{gamback1993pragmatic} and by Ginsberg \cite{ginsberg1999gib}. Amit and Markovitch \cite{amit2006learning} built a decision tree model along with Monte Carlo sampling on top of a human bidding system to resolve the ambiguity of bids.
DeLooze and Downey~\cite{delooze2007bridge} generated examples from a human bidding system, and then used these examples as input for a self-organizing map for ambiguity resolution. In those works, human bidding systems play a central role in AI players' bidding strategy.

A more aggressive route for achieving bridge bidding AI is to teach the AI about bidding without referencing a human bidding system. This was considered by Ho and Lin \cite{ho2015contract} who proposed a decision tree model along with a contextual bandit algorithm. The model demonstrates the possibility to learn to bid directly in a data-driven manner \cite{ho2015contract}. Nevertheless, the study is somewhat constrained by the decision tree model, which comes with a restriction of having at most five choices per bid (decision-tree branch). Moreover, because of the simple linear nodes in the decision tree, the model requires a more sophisticated feature representation of the cards. Ho and Lin \cite{ho2015contract} thus borrowed human knowledge about bidding by encoding the cards as human-designed features, such as the number of cards for each suit. The restrictions on bidding choices and feature representation limit the potential of building a data-driven bidding system for AI.

In all the works discussed thusfar, human-designed features are important either for the human bidding systems within the AI players~\cite{delooze2007bridge}, or for the AI bidding system being learned~\cite{ho2015contract}.
In \cite{ho2015contract}, it was actually reported that raw-card features resulted in considerably worse performance than human-designed features. Inspired by the recent success of deep learning in automatically constructing useful features from raw and abstract ones~\cite{silver2016mastering}, we propose a novel framework
that applies deep reinforcement learning for automatic bridge bidding, which
contributes to advancing the bridge bidding AI in two aspects:

\begin{itemize}
\item learning data representation: Using deep neural networks for feature extraction, our proposed deep reinforcement learning model is the first model that automatically learns the bidding system directly from raw data. Learning from raw data, without relying on human-designed bidding systems or human-designed features, unleashes the full power of machines on using all possible information. The promising performance of our proposed model showcases that learning a bidding system automatically with no human knowledge is possible.
	
\item resolving bid ambiguity: As discussed previously, the main difficulty of bridge bidding is the ambiguity of the bids. Using the proposed reinforcement learning framework, sophisticated bidding rules can be learned automatically to alleviate the ambiguity problem. The reinforcement learning framework arguably mimics what human players do in establishing mutual understanding by practicing together. In this case, however, the framework does so ``together'' \textit{with itself}. To the best of our knowledge, it is the first framework that achieves promising mutual understanding using only machines in bridge bidding.
\end{itemize}

In summary, our proposed deep reinforcement learning framework enables learning complex rules of bidding by nonlinear functions on raw data to avoid ambiguity of the bids and improve bidding performance. In Section 2, we formally establish the problem of bridge bidding as a learning problem. In Section 3, we first introduce reinforcement learning and analyze the key issues in solving the bidding problem. We then propose a novel deep reinforcement learning framework based a modification of Q-learning along with upper-confidence-bound algorithms for balancing between exploration and exploitation. We further introduce a modification of the Bellman's Equation, named penetrative Bellman's Equation, to improve Q-learning. Finally, we discuss our experiments that demonstrate the promising performance of our proposed AI player based on the deep reinforcement learning framework. We demonstrate that the player's bidding performance compares favorably against state-of-the-art AI bidding systems~\cite{ho2015contract} and a contemporary champion-winning bridge software, Wbridge5, which implements a human bidding system.

\section{Problem Setup}

The general bidding problem can be divided into two subproblems, namely bidding without competition, and bidding with competition \cite{ho2015contract}, both of which contain significant amounts of deals in actual bridge games. Bidding without competition assumes that the opponent's team always calls PASS during bidding, and hence information-exchanging would not be blocked; bidding with competition means that both teams want to bid. In this study, we focus on the subproblem of bidding without competition, as with existing works \cite{amit2006learning,delooze2007bridge,ho2015contract}.
The subproblem is a longstanding benchmark for testing computerized bidding systems  \cite{lindelof1983cobra,stanier1975bribip,wasserman1970realization}.
It is also called the bidding challenge, which is common for evaluating the performance of human bridge players during practice sessions and some competitions.

For simplicity, we assume that the bidding team is composed of two
players, Player 1 and Player 2, sitting at the North-South positions,
and their opponents always bid PASS. Player 1 is assumed to bid in
round 1 and the other odd rounds without loss of generality, and
Player 2 is assumed to bid in the even rounds.

We use $\boldsymbol{x}_1$ and $\boldsymbol{x}_2$ to denote the cards of Player~1 and Player 2, and~$\boldsymbol{b}$ to denote the bids of the two players. The element $\boldsymbol{x}_i[k]$ of the boolean array $\boldsymbol{x}_i$ indicates whether Player $i$ holds the $k^{th}$ card in the ordered set of $\{\spadesuit2,\spadesuit3,...\spadesuit A,\heartsuit2,\dots,\heartsuit A,\diamondsuit2,\dots,\diamondsuit A,\clubsuit2,\dots,\clubsuit A\}$. Up to the first $t$ rounds, the element $\boldsymbol{b}^{(t)}[j]$ of the boolean array $\boldsymbol{b}^{(t)}$ indicates whether Player 1 or Player 2 has made the $j^{th}$ bid in the ordered set of $\beta = \{PASS,1\clubsuit,1\diamondsuit,\dots,7NT\}$. As stated in bridge rules, a new bid has to be higher than all previous bids, and thus it is sufficient to infer which bid was bidden by whom given $\boldsymbol{b}^{(t)}$. 

We define the state $s^{(t)}$ to contain $\boldsymbol{x}^{(t)}$ and $\boldsymbol{b}^{(t)}$, where $\boldsymbol{x}^{(t)} = \boldsymbol{x}_1$ for an odd $t$, and $\boldsymbol{x}^{(t)} = \boldsymbol{x}_2$ for an even $t$. The state is defined so that the player bidding in round $t$ can only access her/his own hand. The goal is to find a strategy $G(s^{(t)}) = a^{(t)}$, where the bid $a^{(t)}$ is among the ordered set of $\beta =$ $\{PASS,1\clubsuit,1\diamondsuit,\ldots ,7NT\}$. Note that a further constraint of $ a^{(t)} \geq a^{(t-1)}$ needs to be added to meet the rules of bridge. Furthermore, the ``bid'' of $ a^{(t)} = a^{(t-1)}$ indicates the intent to terminate the bidding procedure, just like the case of $ a^{(t)} = $ PASS. For any given strategy $G$, the array $\boldsymbol{b}^{(t)}$ will be updated by the bid $a^{(t)}$ of the strategy in each bidding round.

We generate the data as follows to learn a good bidding strategy~$G$. For each instance within the data, we directly take the raw card vectors
$(\boldsymbol{x}_1, \boldsymbol{x}_2)$ as the input features. We also generate the score of each possible contract with respect to each $(\boldsymbol{x}_1, \boldsymbol{x}_2)$ to guide learning. Because playing out each possible contract is extremely time-consuming even with computer assistance, the score is calculated
without carrying out the playing phase. In particular, we use double dummy analysis as in previous studies \cite{ho2015contract} to estimate the score for each possible contract.

Double dummy analysis is a technique that attempts to compute the number of tricks taken by each team in the playing phase under perfect information and optimal playing strategy, and is generally considered to be a solved problem in the art of bridge bidding AI. 
Although the analysis is done with an optimistic assumption of perfect information, it has been shown to achieve considerable accuracy with a more rapid analysis than an actual play.


Nevertheless, it is widely known that goodness of bridge bidding is affected by the opponent team's hands. A severe distribution of the opponents' trump cards may cause a good bidden contract to fail. Human players generally bid to maximize the expected score with respect to the opponent team's hands. We approximate the expected score by randomly dealing
the remaining cards to the East and West players for 5 times for each given North-South cards $(\boldsymbol{x}_1,\boldsymbol{x}_2)$. Then, we perform double dummy analysis for each deal, and calculate the final score of each possible contract by averaging on the 5 double dummy analysis results. After obtaining the final score for each possible contract, we store the absolute difference between the final score and the highest final score in a cost vector $\boldsymbol{c}$, where $\boldsymbol{c}[j]$ indicates the penalty of reaching a final contract $j$.

We now formally define our learning problem as follows. Given a data set $D = \{ ( \boldsymbol{ x}_{1n},\boldsymbol{x}_{2n},\boldsymbol{c}_n)\}_{n=1}^N$, where $N$ is the number of instances, we aim to learn a bidding strategy $G$.
For each $(\boldsymbol{x}_1,\boldsymbol{x}_2)$, the strategy $G$ is iteratively fed with the current state $s^{(t)} = \{ ( \boldsymbol{ x}^{{(t)}},\boldsymbol{b}^{(t)})\}$ until it calls PASS or the same bid at some state $s^{(*)}$. The cost of the final bid (contract), namely ${\boldsymbol{c}[G(s^{(*)})]}$, is then used to evaluate $G$. Our goal is to minimize the expected test cost, or equivalently, maximizing the expected test reward of~$G$.

\section{Model}
\begin{figure}[t!]
	\centering
	\includegraphics[width=0.5\textwidth]{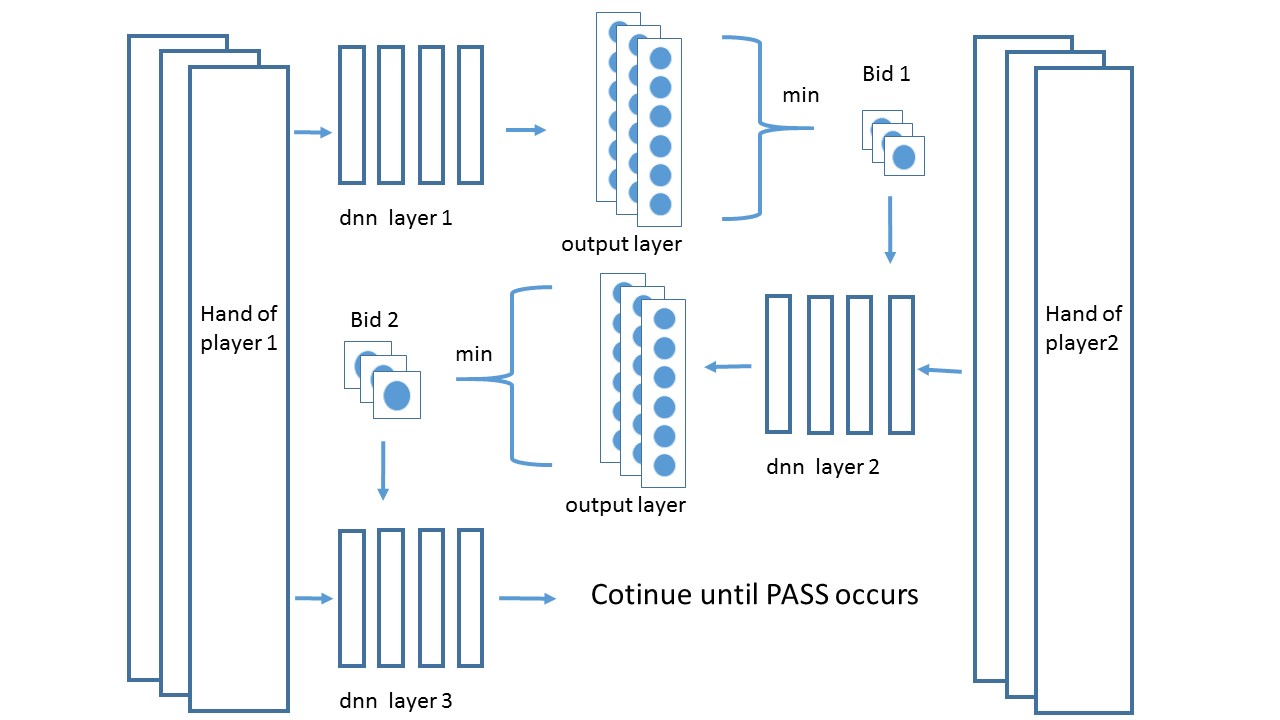}
	\vspace{-2mm}
	\caption{\small The structure of our bridge bidding deep reinforcement learning framework.}
	\vspace{-2mm}
	\label{fig:overview}
\end{figure}
The difficulty of creating a bridge bidding AI lies in that evaluation of a bid is not possible until the end of the playing phase. Even after we introduce double dummy analysis, there is still no obtainable ``score'' until the bidding phase is complete. Therefore, it would be difficult to evaluate bids before the final bid, such as the opening and intermediate bids. It would be most suitable that the intermediate bids are scored by the ability to help the last bid achieve the best score. In this section, we first introduce the reinforcement Q-learning model. Then, we discuss how the model can be extended for conquering the difficulty and solving the bridge bidding problem.

\subsection{Reinforcement Q-Learning}
Q-learning \cite{watkins1992q} is a form of the Reinforcement Learning algorithm that does not require modeling of the environment. In Q-learning, the value function of policies are represented by a two-dimensional lookup table indexed by state-action pairs, defined as $Q(s,a)$. The optimal action-value function $Q^{*}(s,a)$ is defined as the maximum expected return achievable with any strategy after performing an action $a$ in state $s$. Similarly, $Q^{*}(s,\boldsymbol{a})$ is the vector where each value in the vector is obtained by the maximum expected return achievable with any strategy after performing the corresponding action in vector $\boldsymbol{a}$ and in state $s$.

The optimal action-value function follows an important equation known as the Bellman equation. The Bellman equation condiders that, given the current state $s$, all possible actions $a'$, and all resulting states $s'$, the optimal value of $Q^{*}(s, a)$ equals the expected sum of the instant reward $r$ and the total rewards after some best action $a'$ is executed. Formally, we have
\begin{equation}\label{eq:eq1}
 Q^{*}(s,a) = \mathbb{E}_{s'}[ r + \gamma \max_{a'} Q^{*}(s',a') | s,a ] 
\end{equation}

The general idea behind reinforcement learning is to obtain an estimate of the action-value Q-function by continuously modifying it based on feedback from previous actions. After performing an action, the Q-function is updated using the following relation. 
\begin{equation}
Q(s,a) \leftarrow \alpha  Q(s,a)+ (1-\alpha) [r + \gamma \max_{a'} Q^{*}(s',a') | s,a ] 
\end{equation}
The value iteration algorithm converges to the optimal Q-value eventually.
It is important to note that the Q-learning does not specify the next move, and allows arbitrary exploring, because Q-learning constructs a value function on the state-action space, instead of the state space.

For more complicated problems, a non-linear Q-function such as a deep neural network may be used \cite{mnih2013playing,riedmiller2005neural}. Usually an error function is introduced in order to measure the difference between the current and newly experienced Q-values. A Q-network can be trained by minimizing a squared-error measure as demonstrated below,
\begin{equation}
L =  (Q(s,a)-[r + \gamma \max_{a'} Q^{*}(s',a') | s,a ])^2
\end{equation}
Gradient descent techniques such as backpropagation can later be applied to the Q-network in order to minimize the loss function. The update rule is typically applied after each new sample.

\subsection{Modeling for Bridge Bidding}

The bridge bidding problem we consider can been modeled as a two-player partial-information cooperative game with multiple stages. However, the problem of learning a new bidding system can be complicated. It is considerably very difficult to infer the full state by the bidding sequence alone, because each call can either suggest an optional contract or serve as information exchange between partners. A different intention would lead to different inferences under the same situation, thus partners often maintain a list of agreements about the meaning of their bids, widely known as a bidding system. Interestingly, the same exact bid may be considered good in one bidding system and bad in another.

Therefore, while learning a new bidding system, players would need to modify their interpretation of bids alongside their partners. Moreover, the bidding ability of both players may affect the best possible bid in the early stages. For example, if a player is in placing the last bid, he should be more aggressive in his earlier bids. On the other hand, for a player incapable of bidding well in the later bids, it will be better for him to bid more conservatively early on. This signifies that one's bidding competency in the later stage will affect the bidding in the early stage. 

The bidding problem without competition can be viewed as a multi-agent game such that each bidder at different stages of bidding are seen as different players. This model is equivalent to the original bridge bidding problem. In this case, each player has a unique sequence number; players with odd sequence number can share the same 13-card information, while players with even sequence number share another set of 13-card information. Each player knows the bidding result of all the players before him. The game stops when PASS is bid by a player with a sequence number greater than one. Thus, we are able to separate the decision progress of each layer by ``training'' for a different Q-function for each layer of bidding. The algorithm is defined and illustrated in Algorithm \ref{alg:learning}.

In traditional Q-learning, the cost of each bid is updated by Bellman equation as in equation \ref{eq:eq1}. Moreover, the exploration behavior is often demonstrated by an $\epsilon$-greedy strategy that follows the greedy strategy with probability 1- $\epsilon$ and selects a random action with probability $\epsilon$. This forms the baseline algorithm.
\begin{algorithm}[t]
	\label{alg:learning}
	\DontPrintSemicolon
	\caption{The Proposed Learning Algorithm}
	\KwIn{Data = {\{($\boldsymbol{x_{1i}}$,$\boldsymbol{x_{2i}}$,$\boldsymbol{c_{i}}$)\}} 
		$	 \mbox{for } i=1, \ldots, n$
		\\
		Algorithm P to determine cost of action $a$ \\
		Algorithm E to determine exploration and exploitation strategy 
		}
	\KwOut{A bidding strategy G based on the learned $\boldsymbol{\theta_i}$. }
	Initialize action-value function $Q_j$ with random weights $	 \mbox{for } j=1, \ldots, l$
	\\
	\Repeat{enough training iterations by early stopping}
	{
		
		Randomly select a data instance ($\boldsymbol{x_{1i}}$,$\boldsymbol{x_{2i}}$,$\boldsymbol{c_i}$)\\
		\For{round t = 1 to $l$}{
			initialize cost array ${\boldsymbol{c}}(\boldsymbol{a^{(t)}})$\\
			\For{all possible action $a^{(t)}$}{
				determine the cost of action $a^{(t)}$ by P\\
				record resulting cost in ${\boldsymbol{c}}(\boldsymbol{a^{(t)}})$
			}
			save $(S^{(t)},{\boldsymbol{c}}(\boldsymbol{a^{(t)}}))$ in Database D\\
			select action $a^{(t)}$by the highest estimated reward with exploration by E\\
			\uIf{$a^{(t)} == \mbox{PASS}$}
			{
				Break
			}
			update $\boldsymbol{b^{(t+1)}}$ by action $a^{(t)}$  \\
			Set $s^{(t+1)} = (\boldsymbol{x^{(t+1)}},\boldsymbol{b}^{(t+1)})$\\
		}
		\For{round t = 1 to $l$}{
			Sample random minibatch of  $(S^{(t)},{\boldsymbol{c}}(\boldsymbol{a^{(t)}}))$ from D\\
			Perform a gradient descent step on $[(1-{\boldsymbol{c}}(\boldsymbol{a^{(t)}}))-Q(s^{(t)},\boldsymbol{a^{(t)}};\theta)]^2$ }
	}
\end{algorithm}

While Bellman's Equation being a necessary condition for optimality, the convergence time for each Q-function is rather long. Moreover, it has been shown that Q-learning performs considerably poorly in some stochastic environments because of overestimation of action values. In the problem of bridge bidding, being a partial information cooperative game, the overestimation of action values becomes a significant problem.

We define a penetrative Bellman's Equation to deal with the overestimations of action values in the bridge bidding game as follows.
\begin{equation} \label{eq:b1}
Q^{(i)*}(s,a) = \max_{a'} [ Q^{(i+1)*}(s',a') | s,a ] = Q^{(i+1)*}(s^*,a^*| s,a)
 \end{equation}
 where instance reward is zero and $\gamma =1$. We could further apply Bellman's equation on $Q^{(i+1)*}(s^*,a^*)$ as:
\begin{equation}\label{eq:b2}
Q^{(i+1)*}(s^*,a^*) = \max_{a^{(2)}} [   Q^{(i+2)*}(s^{(2)},a^{(2)}) | (s^*,a^*) ]  
\end{equation}
while 
\begin{equation}\label{eq:b3}
\max_{a^{(2)}} [   Q^{(i+2)*}(s^{(2)},a^{(2)}) | (s^*,a^*) ]   =Q^{(i+1)*}(s^{*(2)},a^{*(2)}|(s^*,a^*))
\end{equation}
By combining equation \ref{eq:b1}, equation \ref{eq:b2}, and equation \ref{eq:b3}, we have
\begin{equation} \label{eq:b4}
Q^{(i)*}(s,a) =Q^{(i+2)*}(s^{*(2)},a^{*(2)}|s,a)
\end{equation}
where $s^{*(2)},a^{*(2)}, s^*, a^*,s,$ and $a $ satisfy equation \ref{eq:b1} and equation \ref{eq:b2}.
By recursively applying Bellman's equation, the process stops when $a^{*(t)}$ is the final bid of the game, and thus, the cost of the game can be decided by the precalculated ${\boldsymbol{c}}(a^{*(t)})$. 
\begin{equation} \label{eq:b5}
Q^{(i)*}(s,a) =Q^{(i+t)*}(s^{*(t)},a^{*(t)}|s,a)
\end{equation}
where $s^{*(i)},a^{*(i)}$ represents the best possible action given $s^{*(i-1)},a^{*(i-1)}$ due to the Q-function. Typically, there are less than 6 bids in a game of bridge bidding between the two teammates, therefore the penetrative Bellman's Equation is fairly efficient compared with the original variant, while the cost of each action is much more accurate and this leads to better results, as is discussed in the experiment section later.

Because bridge bidding is a multi-agent cooperative game, the traditional $\epsilon$-greedy algorithm would be detrimental for communication between partners. Note that the main objective in bridge bidding, other than to find the best possible contract, is to convey some information to one's partner. However, randomly bidding any contract in lieu of a certain possibility would make it difficult for the partner of the bidder to understand the current bidding. This will result in poor communication and slower convergence, which has more disadvantages than it has advantages.

Further, exploration is one of the key elements in reinforcement learning to reach the optimum. Without exploration, reinforcement learning will likely be confined to some local optimal because the value of some actions will never be explored. There are various studies investigating the problem of balancing exploration with exploitation. Previous research on exploration of reinforcement learning proposes the use of a sampling technique such as "Thompson sampling" to enhance the performance of exploring\cite{osband2016deep,osband2015bootstrapped}. 

However, most related approaches are not able to deal with one of the key difference between bridge bidding and tradition reinforcement learning: communication with the partner. One of the difficulties of learning a good bidding strategy is the complexity in exploring the value of an action. In games such as chess or go, one may learn that a move is recommended by evaluating the state through playing afterwards. However, in the game of bridge, a bid is only good if one's partner understands the bid and is able to react accordingly. Even in the exploration phase, bids would need to be consistent with the opponent's knowledge. 

Moreover, considering information theory, the exchange of information would work best when the use of each bid is distributed equally. Therefore, we design an exploration scheme using a bandit algorithm. The bandit problem has been a popular research topic in the field of machine learning \cite{auer2002finite,chu2011contextual,langford2008epoch,li2010exploitation}. In the contextual bandit problem, we would like to earn the maximum total rewards within finite tries by pulling a bandit machine from M given machines in a dynamic environment with context. The key is to balance exploration and exploitation. The upper-confidence-bound (UCB) algorithms \cite{chu2011contextual} are some of the most popular contextual bandit algorithms. These algorithms use the uncertainty term to achieve balance. For the bridge bidding problem, we choose to use UCB1\cite{auer2002finite} for its simplicity in connecting with deep neural network and its good performance in previous works.

We now relate the bridge bidding problem to the contextual bandit problem. We assume that each possible bid is a bandit machine, with the context being the cards in one's hand and the previous bids. The reward of each bid is calculated by the penetrative Bellman's Equation, which relates to the final cost vector and future strategy. Nevertheless, there may be uncertainty in terms of the action-value, especially for bids that are rarely used. Contextual bandit can be applied to balance between using the best action inferred by the Q-function and exploring bids that occurs less. The neural network of the Q-function serves as a non-linear version of the reward, the $W^T X$ term in UCB1. Therefore, we formally define algorithm E using UCB1 by selecting $a^{(t)} = \max_{a^{(t)}} [Q(s^{(t)},a^{(t)};\theta)+\alpha  \sqrt{\frac{2\ln{T}}{T_a}}]$, where $T$ is the number of total examples used to learn the entire Q, and $T_a$ is the number of examples such that action $a$ (bid $a$) has been selected. The final algorithm is shown in Algorithm \ref{alg:learning}.

\subsection{Preprocessing and Model Architecture} \label{sec:preprocessing}

The features of training the bridge bidding AIs in previous works include bridge-specific features invented by humans such as high card points\footnote{High Card Points - total points for the picture cards. A=4; K=3; Q=2 ; J=1.}. Deep neural networks are known to contain feature selection function. Therefore, we propose to use 52-dimension raw hand data as our feature. There are debates upon the best form of high card points, since tens and nines may very well serve an important role in certain hands. By using the raw data, we do not limit the computation of the strength of hands by human-designed technique. Moreover, we are able to show that a well-performing bridge bidding system can be designed without human knowledge in bridge.

There are several possible approaches when designing Q-function using a neural network. One approach uses bidding history and the actions as inputs to the neural network
, while another involves listing the cost of all possible outputs, only with the state as input. The drawback of the former is that the computation cost will increase linearly with the number of possible actions. Thus, we choose the latter approach and therefore, it can be stated that the output of the Q-function corresponds to a predicted cost vector of all the possible bids. We denote the action vector $\boldsymbol{a}$ and the true cost vector of all possible bids ${\boldsymbol{c}}(\boldsymbol{a})$. The gradient descent update of the Q-function can be done on $({\boldsymbol{c}}(\boldsymbol{a^{(t)}})-Q(s^{(t)},\boldsymbol{a};\theta))^2$. Notably, there may be actions illegal in certain states because they violate the bridge rule. We set the cost of such actions to an extremely high value so that the rule of bridge can be learned by the Q-function explicitly.

We now present the architecture of the Q-function of the bridge bidding problem. We initialize $l$ separate Q-functions, where $l$ is the length of total bids. For the first Q-function, the input is 52-dimension raw data of Player 1's hand using one-hot encoding, followed by 3 layers of fully connected layers with 128 neurons each. We obtain a 36-dimension output for the cost of each bid. Compared with the first Q-function, in the case of other Q-functions, there is an extra 36-dimension showing the bidding history of the both players, where the 36-dimensions stands for $\{PASS,1\clubsuit,1\diamondsuit,\ldots ,7NT\}$. The bids that have been bidden by any of the two players have the value one, others have the value zero. The final structure of our learning framework is shown in Figure \ref{fig:overview}.

\section{Experiment}
\begin{table*}[t!]
	\centering
	\caption{ Comparisons between the average cost of two exploration methods where 3 values of parameter in each exploration method is tested}
	\label{tableexplore}
	\begin{tabular}{l l l l l}
		\hline
		layer                               & 2 & 3 & 4 & 5 \\ \hline
		$\epsilon$-Greedy, $\epsilon$ = 0.001 & 2.9628 $\pm$0.0257 & 2.7748 $\pm$0.0328 & 2.7648 $\pm$0.0290& 2.7650 $\pm$0.0031\\ \hline		
		$\epsilon$-Greedy, $\epsilon$ = 0.005 & 2.9582 $\pm$0.0036 & 2.8201 $\pm$0.0282 & 2.7773 $\pm$0.0419& 2.7510 $\pm$0.0457\\ \hline		
		$\epsilon$-Greedy, $\epsilon$ = 0.01 & 2.9857 $\pm$0.0227 & 2.8080 $\pm$0.0113 & 2.7696 $\pm$0.0179& 2.7716 $\pm$0.0305\\ \hline
		$\epsilon$-Greedy, $\epsilon$ = 0.05 & 3.0125 $\pm$0.0689 & 2.8331 $\pm$0.0413& 2.8408 $\pm$0.0367& 2.8328 $\pm$0.0079\\ \hline
		$\epsilon$-Greedy, $\epsilon$ = 0.1  & 3.0575 $\pm$0.0092& 2.8758 $\pm$0.0240 & 2.8679 $\pm$0.0228& 3.0035 $\pm$0.1720\\ \hline
		no exploration                       & 2.9600  $\pm$0.0372& 2.7914 $\pm$0.0080& 2.7559 $\pm$0.0616& 2.7949 $\pm$0.0358\\ \hline
		UCB1, $\alpha$ = 0.05                & 2.9329 $\pm$0.0069& 2.7776 $\pm$0.0128 & 2.7451 $\pm$0.0055& 2.7695 $\pm$0.0143\\ \hline
		UCB1, $\alpha$ = 0.1                 & 2.9391 $\pm$0.0279 & 2.7289 $\pm$0.0595 & 2.6984 $\pm$0.0207& 2.7397 $\pm$0.0187\\ \hline
		UCB1, $\alpha$ = 0.2                 & 2.9542  $\pm$0.0052& 2.8042 $\pm$0.0358& 2.7183 $\pm$0.0171& 2.7465 $\pm$0.0221\\ \hline
		
	\end{tabular}
\end{table*}
We compare the proposed model with a baseline model, a state-of-the-art model, and a well-known computer bridge software, Wbridge5 \cite{costel2014wbridge5}, which has won the computer bridge championship for several years. We randomly generate a dataset of 140,000 instances to be used in our experiments. We use 100,000 instances for training, and split the other 40,000 instances evenly for validation and testing. We compare the sparse binary features for representing the existence of each card to the condensed feature with second order extension used in previous works~\cite{ho2015contract}.

We set the cost vectors ${\boldsymbol{c}}(\boldsymbol{a})$ from International Match Points, which is an integer between 0 and 24 that is commonly used for comparing the relative performance of two teams in most bridge game. The cost vector is obtained by subtracting the cost of action array by the best possible bid followed by a normalization step, that is, ${\boldsymbol{c}}(\boldsymbol{a}) = [{\boldsymbol{c}}(\boldsymbol{a})' - min_a {\boldsymbol{c}}(a)']/25$. The result is divided by 25 to ensure the cost is normalized between 0 and 1. ${\boldsymbol{c}}(\boldsymbol{a})'$ denotes the origin cost of each action calculated by the double dummy analysis\footnote{One technical detail is that ${\boldsymbol{c}}$ is generated by assuming that the player who can win more tricks in the contract is the declarer}. The cost can be transformed to the reward using $R(\boldsymbol{a}) = 1-{\boldsymbol{c}}(\boldsymbol{a})$. We set the cost of the rule-breaking bids to 1.2, therefore letting the bidding system learn the bidding rules implicitly. Moreover, the bids in the testing phase are chosen from legal bids.

For deep neural networks, rmsprop is used to speed up the convergence time. In the following experiments, we fix the parameters related to the deep neural network. The following parameters were used in the experiments of the fully connected deep neural network: decay = 0.98, momentum = 0.82, step rate = 0.83, batchsize = 50 and $\eta$ = 0.05. These parameters remain unchanged during our experiments because the focus of our study is not on deep neural network parameters. We use early stopping from the validation result to determine the number of epochs to end the training.

\subsection{Exploration Method}

We compare the two exploration model, $\epsilon$-greedy exploration and UCB1, for the exploration algorithm E in algorithm 1. The parameter in $\epsilon$-greedy exploration is to choose a random action with possibility $\epsilon$ and follow the best action given the Q-function otherwise. The parameter in UCB1 exploration is to select $a^{(t)} = {max}_{a^{(t)}} [Q(s^{(t)},a^{(t)};\theta)+\alpha  \sqrt{\frac{2\ln{T}}{T_a}}]$, where T is the number of total examples used to learn the complete Q, and $T_a$ is the number of examples such that action $a$ (bid $a$) has been selected. We perform experiments on $\epsilon \in \{0.001,0.005, 0.01, 0.05, 0.1\}$ and $\alpha \in \{0.05, 0.1, 0.2\}$. We also list the result where no exploring methods are used.

We can see from the table that the UCB1 exploration generally outperforms that by $\epsilon$-greedy, showing that the UCB1 exploration fits well with the model. The $\epsilon$-greedy exploration method performs even worse than in the case of no exploration for $\epsilon \geq$ 0.05, which is arguably because of the enhanced ambiguity of random exploration. It is noteworthy that the no exploration method does include some sense of exploration in the dnn structure itself, because all the possible actions are updated in certain Q-functions, thus they contain the actions which are not likely to be chosen as well. However, deep exploration such as in the case of UCB1 can further improve the result as shown in Table \ref{tableexplore}. The best parameter for UCB1 exploration is $\alpha$ = 0.1 for layer $\leq$ 3, and $\alpha$ = 0.05 for layer $ = $ 2. These parameter values will be used in the experiments discussed in the remainder of this paper.

\begin{table}\centering
	\caption{ Comparisons between different methods of updating the Q-functions}
	\label{bellman}
	
	\begin{tabular}{lllll}
		\hline
		Total bids                          & 2      & 3     & 4    & 5    \\ \hline
		Baseline                       & 2.9308 & 2.8585 & 2.8795 & 2.9225 \\ 
		Penetrative Bellman's Equation & 2.9329   & 2.7289 & 2.6984 & 2.7397 \\
		\hline
	\end{tabular}
\end{table}

\subsection{Penetrative Bellman's Equation}

For the baseline model, we use Bellman's Equation and UCB1 exploration as Algorithm P and Algorithm E in Algorithm 1. We compare the result of Bellman's Equation and penetrative Bellman's Equation for candidates of algorithm P.

In Table \ref{bellman}, the average test cost is shown with the total bids as variables. UCB1 is used as exploration method with the exploration parameter $\alpha$ set to 0.1. We can see that when the total bids are larger than 2, Penetrative Bellman's Equation outperforms the baseline method by a considerable margin. In fact, the baseline model has little variance of performance with varying total bids. This can be attributed to the cumulation of estimation errors of the Q-function when the total bids increase. Therefore, the performance improvement of the complexity of the model is cancelled out by the cumulated error. We can infer that the primary reason that Penetrative Bellman's Equation is effective is that it enables the possibility to learn a deep Q-learning model with more total bids, by providing a more accurate estimation of cost.
\begin{table}\centering
	\caption{ Comparisons of average cost between different models and \cite{ho2015contract} and wbridge5}
	\label{traintestvali}
	\begin{tabular}{lllll}
		
		Model                         & training     & validation     & testing     \\ \hline
		layer = 2  & 2.8013 $\pm$ 0.0368 & 2.9150 $\pm$0.0049 & 2.9329 $\pm$ 0.0069\\ 
		layer = 3  & 2.6725 $\pm$ 0.0392 & 2.7363 $\pm$0.0465 & 2.7289 $\pm$ 0.0595\\ 
		layer = 4  & 2.5992 $\pm$ 0.0474 & 2.6700 $\pm$0.0245 & 2.6984 $\pm$ 0.0207\\ 
		layer = 5  & 2.6442 $\pm$ 0.0261 & 2.7150 $\pm$0.0123 & 2.7397 $\pm$ 0.0187\\  \hline
		\cite{ho2015contract} layer = 4  & 2.9730 $\pm$ 0.0315 & 3.0697 $\pm$0.0388 & 3.0886 $\pm$ 0.0479\\ 
		\cite{ho2015contract} layer = 6 & 2.9136 $\pm$ 0.0384 & 3.1267 $\pm$0.0092 & 3.1657 $\pm$ 0.0199\\ 
		Wbridge5 & N/A   & N/A & 3.0039 \\ \hline
		\hline
	\end{tabular}
\end{table}

\begin{table*}[t]
	\centering
	\caption{5 examples where features of Player 1 is listed in the actual column, and the estimation from Player 2 is listed in the estimate column. The bidding history along with the best bid and cost are listed in the table too.}
	\label{exampletable}
	\begin{tabular}{|l|l|l|l|l|l|l|l|l|l|l|}
		\hline
		& actual         & estimate        & actual         & estimate         & actual         & estimate         & actual         & estimate        & actual        & estimate        \\ \hline
		number of spades           & 5              & 5.1024          & 2              & 2.2202           & 3              & 3.1463           & 4              & 4.5931          & 3             & 2.7333          \\
		number of hearts           & 2              & 2.2983          & 4              & 4.9368           & 4              & 3.5169           & 3              & 2.5334          & 1             & 1.0620           \\
		number of diamonds         & 3              & 2.3366          & 5              & 2.9966           & 4              & 3.8936           & 2              & 3.0599          & 6             & 5.9515          \\
		number of clubs            & 3              & 3.2061          & 2              & 2.9111           & 2              & 3.3676           & 4              & 2.8925          & 3             & 3.2824          \\
		HCP             & 5              & 4.9745          & 21             & 19.5970           & 9              & 9.7680            & 19             & 18.5185         & 5             & 5.7930           \\ \hline
		bidding history & \multicolumn{2}{l|}{$P$-$1NT$-$2\spadesuit$-$4\heartsuit$} & \multicolumn{2}{l|}{$1\spadesuit$-$1NT$-$3\spadesuit$-$4\heartsuit$} & \multicolumn{2}{l|}{$P$-$1\clubsuit$-$1NT$-$1NT$} & \multicolumn{2}{l|}{$1\spadesuit$-$2\clubsuit$-$5\heartsuit$-$6\heartsuit$} & \multicolumn{2}{l|}{$P$-$1\heartsuit$-$2\diamondsuit$-$2\diamondsuit$} \\
		best contract       & \multicolumn{2}{l|}{$4\heartsuit$}          & \multicolumn{2}{l|}{$4\heartsuit$ or $3NT$}    & \multicolumn{2}{l|}{$2\diamondsuit$}           & \multicolumn{2}{l|}{$7NT$ or $7\heartsuit$}   & \multicolumn{2}{l|}{$2\diamondsuit$ or $3\diamondsuit$}   \\
		cost(IMP)       & \multicolumn{2}{l|}{0}           & \multicolumn{2}{l|}{0}            & \multicolumn{2}{l|}{4}            & \multicolumn{2}{l|}{11}          & \multicolumn{2}{l|}{0}          \\ \hline
	\end{tabular}
\end{table*}

\subsection{Comparison with the State-of-the-Art}
We now discuss the experiment with different model structures. We consider the Q-learning model with total bids $\in \{2,3,4,5\}$. For each model structures, we use the validation result to determine the parameter $\alpha$, where $\alpha \in \{0.05,0.1,0.2\}$. We compare the bridge bidding result with that in the work of \cite{ho2015contract}, and a well-known computer bridge software, Wbridge5 \cite{costel2014wbridge5}, which has won the computer bridge championship for several years. It is known that Wbridge5 adopts some human bidding conventions. It is believed that wbridge5 uses a Monte-Carlo search in the bidding process, so it takes a considerable time to make a bid. The result will be influenced by potential limits on bidding time. On the other hand, despite the longe training time of deep neural network, our bridge bidding model is able to decide the bid instantaneously. We run the same 140,000 data on our model and that proposed \cite{ho2015contract}, while running the 20,000 testing data on Wbridge5, using the code provided by the author of \cite{ho2015contract}.

The results in Table 3 shows that the deep reinforcement learning model with layer = 4 has the best performance among all models. Moreover, each deep reinforcement learning model outperforms the result reached by Wbridge5. This showcases that deep reinforcement learning achieves a good result even with a simple model structure. The result justifies that there is indeed a considerable potential to improve the traditional approach of bridge bidding AI by "borrowing" human bidding systems.

\subsection{Computational Time}

In this section, we discuss the computing time for training our models. The code is written in MATLAB and executed on a Ubuntu Linux 12.04 LTS AMD64 system, using Intel Xeon X5560 CPU with 60 GB RAM. We list the training time for one epochs and the total training time with the model with different layers.
\begin{table}
	\centering
	\caption{approximate computation time for each model}
	\label{compute}
	\begin{tabular}{lllll}
		layer                             & 2   & 3   & 4   & 5    \\ \hline
		running time per epoch (sec)      & 121 & 278 & 492 & 713  \\
		running time until converge (hrs) & 0.5 & 2.3 & 6.8 & 17.9
		\\ \hline
	\end{tabular}
\end{table}
In Table \ref{compute}, we can observe that the training for models of 2 and 3 layers is quite efficiency, whereas the training time becomes considerably long for larger models. The total converging time is approximately in the order of $l^3$, where l is the total layer (or total bid) in the model. This is because the complexity of penetrative Bellman's Equation has an extra order of $l$ compared with the complexity of Bellman's Equation.

\subsection{Understanding the Bidding System}
One may wonder by learning a new bidding system, is the result of new bidding system understandable? In this section, we show that the learned bidding system is understandable by the bidder's partner, and show the opening table of our bidding system.

We design an experiment testing whether the feature of the partner's hand will be learned along the deep learning training process. In section \ref{sec:preprocessing}, we define the output layer of the dnn as a 36 dimension vector containing the cost of each possible bid. We add a 5 dimension vector to the output layer representing the number of cards in each suit and the high card points, which is usually used as the deciding feature of human bidding. That is, we let our bidding algorithm learn a 5- dimension representative of the partner's hand along with the bidding system. We use the model with total bid = 4 and $\alpha$ = 0.1 in the experiment.
\begin{table}\centering 
	\caption{The resulting average cost with and without the extra 5 dimension containing partner's hand information in the training objective, where ours* contains the extra 5 dimension}
	\label{twoitems}
	\begin{tabular}{lllll}
		
		Model    &  Ours  & Ours*      \\ \hline
		Cost  &  2.6984 $\pm$ 0.0207& 2.6910 $\pm$ 0.0275   \\ 
		\hline 
	\end{tabular}
\end{table}
The results in Table \ref{twoitems} indicate whether adding the extra 5-dimension vector to the training objective leads to a considerable difference in the average cost. In order to give a more concrete idea of the result, we randomly choose 5 examples in the testing data where the bidding remains for at least 4 turns. After Player 2 received the third bid (which is bid by Player 1), we compare the estimation of the feature of Player 1 by Player 2 with the actual feature of Player 1. Thes result is shown in Table \ref{exampletable}. The result demonstrates that Player 2 is able to precisely estimate the hand of Player 1, even when Player 1 has only made two bids.

We list the opening table for the model trained in Table \ref{opentable}. The result is compared with SAYC and \cite{ho2015contract}. The abbreviation "bal" refers to a balanced distribution of cards in each suits.
\begin{table}
	\centering
	\caption{Opening Table comparison}
	\label{opentable}
	\begin{tabular}{llll}
		Bid  & ours             & SAYC                & {[}1{]}                     \\ \hline
		PASS & 0-10 HCP         & 0-11 HCP            & 0-12 HCP                    \\
		$1\clubsuit$   & 11+ HCP          & 12+ HCP, 3+$\clubsuit$          & 9-19 HCP, 4-6 $\heartsuit$             \\
		$1\diamondsuit$   & 10+ HCP, 5+$\heartsuit$      & 12+ HCP, 3+$\diamondsuit$         & 8-18 HCP, short$\spadesuit$   and 4-6$\clubsuit$  \\
		$1\heartsuit$   & 12+ HCP, 5+$\spadesuit$       & 12+ HCP, 5+$\heartsuit$         & 12-23 HCP, w/o long suit    \\
		$1\spadesuit$   & 16+HCP, bal & 12+ HCP, 5+$\spadesuit$         & 10-19 HCP, 4-6 $\spadesuit$              \\
		$1NT$  & 12+ HCP, 6+$\diamondsuit$      & 15-17 HCP, bal & Not used                    \\
		$2\clubsuit$   & Not used         & 22+ HCP             & 0-17 HCP, long $\clubsuit$            \\
		$2\diamondsuit$   & Not used         & 5-11 HCP, 6+$\diamondsuit$        & 0-17 HCP, long $\diamondsuit$             \\
		$2\heartsuit$   & 18+ HCP, 5-6$\spadesuit$      & 5-11 HCP, 6+$\heartsuit$        & 0-13 HCP, long $\heartsuit$             \\
		$2\spadesuit$   & Not used         & 5-11 HCP, 6+$\spadesuit$         & 0-13 HCP, long $\spadesuit$              \\
		$2NT$  & 15-17 HCP, 6+$\spadesuit$       & 20-21 HCP, bal & Not used                   
	\end{tabular}
\end{table}

\subsection{Different Initialization of Opening Bid}
Thus far, this paper has been focusing on learning the bidding system without the aid of current human bidding system. In this section, we discuss an initial attempt to study the possibility to use deep reinforcement learning to enhance the current human bidding system. There is an opening bid for each human bidding system; a well-known bidding system is the Standard American Yellow Card (SAYC). Th bidding system is used widely by both amateur and professional players. In this section, we implement a structure combining the SAYC open bid and our deep reinforcement learning algorithm by fixing the first bid with a written open table of SAYC. Therefore, our learning algorithm learns all the bids after the opening bid, where the opening bid is fixed. There is a bid called weak bid in SAYC opening serving the purpose of interfering with opponents ($2\diamondsuit$,$2\heartsuit$,$2\spadesuit$ in Table \ref{opentable}). Because there is no need of a weak bid in this subproblem of bidding with no competitions, we perform experiments on both SAYC opening with and without weak bidding, which is denoted as ours-SAYC and ours-SAYCNW. We run experiment with our best model, where the total bid is 4 and UCB1 $\alpha$ 0.1. The result is compared with Wbridge5 as a comparison baseline.

\begin{table}\centering
	\caption{ Comparisons of average cost between initialization models}
	\label{withinitial}
	\begin{tabular}{lllll}
		
		Model    & Wbridge5 & Ours  & Ours - SAYC     & Ours - SAYCNW   \\ \hline
		Cost  & 3.0039 & 2.6793 & 2.8004 & 2.7795  \\ 
		\hline \hline
	\end{tabular}
\end{table}
We can see that in Table \ref{withinitial}, our model with SAYC initialization for opening bid outperforms the result of Wbridge5 by a considerable margin. This verifies that our bidding model is able to not only learn well, but also is and improvement on the existing human bidding system. Moreover, our model has a lower cost than the model with SAYC initialization, hinting that using human bidding system may be limiting the potential power of computer bridge bidding.

\section{Conclusion and Future Works}
We propose a novel model that automatically learns to bid from raw hand data by coupling deep reinforcement learning with improved exploration and update techniques. To the best of our knowledge, our proposed model is the first to tackle automatic bridge bidding from raw data without additional human knowledge.
We demonstrate that our proposed model outperforms champion-winning programs and state-of-the-art models by a considerable margin. The superior performance validates the potential of deep learning for reaching a competitive bidding system on its own.

We believe that it is possible to extend our model for the other
subproblem: bidding with competition. In particular, the flexibility of the proposed model allows it to improve its bidding strategy, with or without competition, by self-playing as its own opponent team, or by playing with other human or AI teams.



\ack
We thank the anonymous reviewers for valuable suggestions. We also thank Chun-Yen Ho and other members of the NTU CLLab for insightful discussions. This material is based upon work supported by the Air Force Office of Scientific Research, Asian Office of Aerospace Research and Development (AOARD) under award number FA2386-15-1-4012, and by the Ministry of Science and Technology of Taiwan under number MOST 103-2221-E-002-148-MY3.

\bibliographystyle{ecai}
\bibliography{ecai}
\end{document}